\documentclass[pdflatex,sn-basic,Numbered,iicol]{sn-jnl}

\usepackage{graphicx}
\usepackage{multirow}
\usepackage{amsmath,amssymb,amsfonts}
\usepackage{amsthm}
\usepackage{mathrsfs}
\usepackage[title]{appendix}
\usepackage{xcolor}
\usepackage{textcomp}
\usepackage{manyfoot}
\usepackage{booktabs}
\usepackage{algorithm}
\usepackage{algorithmicx}
\usepackage{algpseudocode}
\usepackage{listings}

\theoremstyle{thmstyleone}

\theoremstyle{thmstyletwo}

\theoremstyle{thmstylethree}

\begin{document}

\title[GNN for Social Network Influence]{Graph Neural Networks for Influence Maximization in Social Networks: An Unsupervised Minimum Dominating Set Approach}

\author*[1]{\fnm{Erfan} \sur{Ahmadi}}\email{e.ahmadi@teias.institute}

\author[2]{\fnm{Mina} \sur{Shirazi}}\email{minashirazi@ut.ac.ir}

\author[1]{\fnm{Behnam} \sur{Bahrak}}\email{b.bahrak@teias.institute}

\affil*[1]{\orgdiv{Computer Engineering Department}, \orgname{Tehran Institute for Advanced Studies (TeIAS)}, \orgaddress{\city{Tehran}, \country{Iran}}}

\affil[2]{\orgname{University of Tehran}, \orgaddress{\city{Tehran}, \country{Iran}}}

\abstract{The Minimum Dominating Set (MDS) problem is a classic NP-hard combinatorial optimization problem with critical applications in social network analysis, including viral marketing, influence maximization, public health interventions, and information dissemination. Identifying a minimal set of influential individuals whose reach covers an entire social network is central to these applications, yet remains computationally challenging at scale. Graph neural networks (GNNs) have emerged as powerful tools for learning over graphs, and recent work explores their application to hard combinatorial problems. This paper presents a novel unsupervised GNN framework for the MDS problem that eliminates the need for ground-truth solutions during training. Trained on 12,000 synthetic graphs with diverse structural properties, our method achieves up to 55$\times$ faster inference than metaheuristic baselines and up to 14$\times$ faster inference than supervised learning approaches, while finding optimal or near-optimal dominating sets on real-world social network benchmarks. Our learned heuristic generalizes effectively to unseen graph distributions, demonstrating strong practical applicability for large-scale social network analysis.}

\keywords{Graph neural networks, dominating set, combinatorial optimization, social networks, influence maximization, unsupervised learning}

\maketitle

\section{Introduction}\label{sec:intro}

Dominating set is one of the central problems in graph theory and combinatorial optimization. Given a graph $G = (V, E)$, a dominating set is a subset $D \subseteq V$ such that every vertex is either in $D$ or adjacent to a node in $D$. The classical Minimum Dominating Set (MDS) problem asks for a dominating set of minimum size. It is NP-complete and belongs to the family of problems used by Karp in his foundational work on NP-completeness~\cite{karp1972reducibility}. Even approximating MDS is hard: its approximation behavior is closely tied to the well-known hardness of set cover, where Feige showed a threshold of $(1 - o(1)) \ln n$ for polynomial-time approximation~\cite{feige1998setcover}.

Traditionally, MDS is tackled with greedy approximation algorithms or combinatorial heuristics. Greedy set cover yields an $O(\ln \Delta)$-approximation (where $\Delta$ is the maximum degree), and more refined methods exploit structural properties of specific graph families. However, in many modern applications---such as large social networks, wireless sensor networks, or biological interaction networks---graphs are huge and noisy, and classical heuristics may be suboptimal or require careful hand-tuning.

The MDS problem has particularly important applications in online social networks, where it addresses fundamental questions in viral marketing and influence maximization. In viral marketing, companies seek to identify a minimal set of influential individuals to target with promotional content, such that information cascades through friend connections to reach the entire network~\cite{kempe2003influence}. The Positive Influence Dominating Set (PIDS) variant, where each person must have a sufficient fraction of their friends in the target set to be influenced~\cite{wang2009pids}, models realistic social influence dynamics better than classical domination.

These applications extend to diverse domains: political campaigns identifying key voters to reach entire communities, public health interventions selecting individuals for behavioral change programs (addressing issues like smoking cessation or healthy lifestyle adoption), and information dissemination strategies for emergency alerts or rumor blocking. Scale-free social networks, which follow power-law degree distributions, present unique challenges as dominating sets in these networks tend to be larger than in random graphs~\cite{molnar2013minimum}, making efficient heuristics essential for practical deployment.

Over roughly the last decade, there has been growing interest in \emph{neural combinatorial optimization}, where deep networks are used to learn heuristics for NP-hard problems~\cite{vinyals2015pointer,kool2019attention,dai2017learning}. In parallel, graph neural networks (GNNs) have become a dominant paradigm for learning over graph-structured data~\cite{velickovic2018gat}. Several recent works explore GNN-based approaches for classical graph problems, including vertex cover, max-cut, and dominating set.

Our approach offers several key advantages over existing methods:
\begin{itemize}
    \item \textbf{Unsupervised training:} Our framework requires no labeled solutions or ground-truth dominating sets during training, making it practical for real-world scenarios where optimal solutions are computationally expensive to obtain.
    \item \textbf{Novel probabilistic loss:} We introduce a multi-objective loss function combining three carefully designed components that provide stronger training signals than previous approaches, yielding near-optimal dominating sets without supervision.
    \item \textbf{Strong generalization:} Our model successfully scales to graphs larger than those seen during training, maintaining near-optimal performance across diverse graph structures and distributions including real-world social networks.
    \item \textbf{Computational efficiency:} Following the Global Prediction (GP) paradigm, the GNN produces node-wise probabilities in a single forward pass, offering significant speed advantages over sequential construction methods while maintaining solution quality through our ensemble decoding approach.
\end{itemize}

The structure of this paper is as follows. We first present a comprehensive review of related work in Section~\ref{sec:related}, covering classical algorithms, neural approaches to combinatorial optimization, and recent GNN-based methods for graph problems. In Section~\ref{sec:methodology}, we detail our proposed methodology, including the GNN architecture, the probabilistic loss formulation, and the multi-strategy decoding algorithm. Section~\ref{sec:applications} discusses practical applications and addresses architectural considerations. Section~\ref{sec:experimental} presents our experimental setup and results. Finally, we conclude with a discussion of implications and future directions.

\section{Related Work}\label{sec:related}

\subsection{Classical Algorithms and Theory for MDS}

The Minimum Dominating Set problem has been extensively studied in the algorithms and complexity literature. The connection between MDS and set cover establishes that a simple greedy algorithm achieves an $O(\ln \Delta)$-approximation ratio, where $\Delta$ is the maximum degree~\cite{karp1972reducibility}. This bound is essentially tight under standard complexity assumptions, as Feige demonstrated that approximating set cover (and hence MDS) within $(1-o(1))\ln n$ is NP-hard~\cite{feige1998setcover}.

For special graph classes, better approximation ratios or exact polynomial-time algorithms exist. Trees admit linear-time exact algorithms via dynamic programming, and planar graphs allow polynomial-time approximation schemes. Various dominating set variants have been studied, including $k$-dominating sets where each non-dominated vertex must have at least $k$ neighbors in the dominating set~\cite{nguyen2020kdominating}, connected dominating sets where the induced subgraph must be connected~\cite{butenko2003cds,wan2003cds}, and positive influence dominating sets motivated by social network applications~\cite{dinh2011pids,raghavan2022pids}. These problems find applications in wireless sensor networks, power grid monitoring~\cite{haynes2002power}, and biological network analysis~\cite{milenkovic2011dominating}.

\subsection{Neural Approaches to Combinatorial Optimization}

The emergence of deep learning has motivated researchers to explore neural methods for solving combinatorial optimization problems. Vinyals et al.\ introduced Pointer Networks, a sequence-to-sequence architecture with attention mechanisms for combinatorial problems with variable-size outputs~\cite{vinyals2015pointer}. This work demonstrated that neural networks could learn solution strategies for problems like the Traveling Salesman Problem without explicit algorithmic design. Kool et al.\ extended this approach with the Attention Model, showing that pure attention-based architectures could achieve competitive performance on routing problems~\cite{kool2019attention}.

For graph-structured problems specifically, Khalil et al.\ proposed a reinforcement learning framework that combines graph embeddings with Q-learning to tackle problems such as vertex cover and maximum clique~\cite{dai2017learning}. Their Structure2Vec approach learns node embeddings that capture local graph structure and uses these embeddings to guide a greedy construction procedure. This work established the feasibility of learning-based heuristics for NP-hard graph problems and demonstrated generalization to larger problem instances than those seen during training.

\subsection{GNN-Based Methods for Graph Problems}

Graph Neural Networks have become the dominant architecture for learning on graph-structured data~\cite{velickovic2018gat}. Recent work has explored using GNNs specifically for combinatorial optimization on graphs. Karalias and Loukas introduced the Erdős Goes Neural framework, an unsupervised learning approach for graph combinatorial optimization~\cite{karalias2020erdos}. Their method trains GNNs to produce node scores that can be decoded into solutions for problems like maximum independent set and minimum vertex cover, using differentiable approximations of the discrete optimization objectives.

Wenkel et al.\ proposed GCON, a multi-filter GNN architecture designed for combinatorial optimization~\cite{wenkel2025gcon}. They demonstrate that using multiple graph filters with different spectral properties enables the network to capture both local and global graph structure, leading to improved performance on various optimization problems. Their theoretical analysis connects the approximation capabilities of GNNs to classical algorithms, showing that certain GNN architectures can provably achieve competitive approximation ratios.

From a theoretical perspective, Sato et al.\ analyzed the approximation ratios achievable by GNNs for combinatorial problems~\cite{sato2019approxgnn}. They established both positive results (showing that certain GNN architectures can match classical approximation guarantees) and negative results (identifying structural limitations that prevent GNNs from achieving arbitrarily good approximations for some problems). This work provides important theoretical grounding for understanding when and why GNN-based approaches can succeed on combinatorial optimization tasks.

Most directly relevant to our work, Kothapalli et al.\ developed a learning-based heuristic specifically for the minimum dominating set problem using Graph Convolutional Networks~\cite{kothapalli2023mds}. They train a GCN to score nodes for inclusion in the dominating set and integrate these scores into an iterative greedy algorithm. Their evaluation on synthetic and real-world graphs demonstrated improvements over classical greedy baselines. Our work builds on this foundation but introduces a novel unsupervised loss formulation with explicit coverage and size penalties, eliminates the need for ground-truth solutions during training, and incorporates an equalizer term that encourages more decisive probability predictions.

\subsection{Machine Learning Paradigms for Combinatorial Optimization}

Recent work has categorized machine learning approaches for combinatorial optimization into three main paradigms based on how solutions are constructed: Global Prediction (GP), Local Construction (LC), and Adaptive Expansion (AE)~\cite{ma2025coexpander}.

\emph{Global Prediction (GP)} models make a single forward pass through the neural network to predict probabilities or scores for all solution elements simultaneously, then use a deterministic or randomized decoding procedure to extract a valid solution. This one-shot approach offers computational efficiency but requires the model to capture the entire solution structure in a single prediction. Our approach falls into this category.

\emph{Local Construction (LC)} models build solutions incrementally through sequential decision-making, typically using reinforcement learning. At each step, the model selects the next element to add based on the current partial solution. While often more accurate, LC methods require multiple forward passes (one per decision).

\emph{Adaptive Expansion (AE)} models combine aspects of both paradigms, making initial global predictions that are iteratively refined through multiple passes. Examples include diffusion-based solvers and certain evolutionary approaches.

For the MDS problem, GP approaches are particularly attractive due to their computational efficiency and ability to leverage unsupervised training objectives that directly encode problem constraints.

\section{Methodology}\label{sec:methodology}

We present a graph neural network-based framework for solving the Minimum Dominating Set problem in an unsupervised manner. Our approach consists of three main components: a message-passing GNN architecture that produces node-wise probabilities, a probabilistic loss function that approximates the discrete MDS objective, and a multi-strategy decoding algorithm that converts probabilities into concrete dominating sets.

\subsection{Graph Neural Network Architecture}

Our model employs a two-layer Graph Convolutional Network (GCN) as the core 
architecture. We also evaluated a Graph Attention Network (GAT) variant under 
the same training pipeline; it consistently underperformed the GCN-based model 
in preliminary experiments, so we adopt GCN for all results reported in this 
paper. For a graph $G = (V, E)$ with $n = |V|$ vertices, we construct initial node features by concatenating two components: a binary indicator vector representing initial vertex properties and a degree-based feature vector. These features are denoted as $\mathbf{X} \in \mathbb{R}^{n \times d}$ where $d=2$ in our base configuration.

For the GCN variant, each layer applies the transformation:
\[
\mathbf{H}^{(l+1)} = \sigma\left(\tilde{\mathbf{D}}^{-1/2}\tilde{\mathbf{A}}\tilde{\mathbf{D}}^{-1/2}\mathbf{H}^{(l)}\mathbf{W}^{(l)}\right)
\]
where $\tilde{\mathbf{A}} = \mathbf{A} + \mathbf{I}$ is the adjacency matrix with added self-loops, $\tilde{\mathbf{D}}$ is the corresponding degree matrix, $\mathbf{W}^{(l)}$ are learnable weight matrices, and $\sigma$ is a ReLU activation function. The initial features $\mathbf{H}^{(0)} = \mathbf{X}$ are propagated through two GCN layers with hidden dimension $h=32$ and dropout rate $p=0.2$ for regularization.

After the final GCN layer, we obtain node embeddings $\mathbf{V}_n \in \mathbb{R}^{n \times h}$. To predict the probability that each node should be included in the dominating set, we concatenate each node's embedding with its degree information and pass the result through a three-layer MLP:
\[
p_i = \sigma\left(\text{MLP}([\mathbf{v}_i; d_i])\right)
\]
where $\mathbf{v}_i$ is the embedding for node $i$, $d_i$ is its degree, and $\sigma$ is the sigmoid activation function, producing probabilities $p_i \in (0,1)$.
\subsection{Probabilistic Loss Function}

The key innovation in our approach is an unsupervised loss function that directly optimizes the trade-off between minimizing the size of the dominating set and ensuring complete coverage of all vertices. The total loss consists of three weighted terms:
\[
\mathcal{L}_{\text{total}} = \alpha \cdot \mathcal{L}_{\text{size}} + \beta \cdot \mathcal{L}_{\text{coverage}} + \gamma \cdot \mathcal{L}_{\text{equalizer}}
\]

\paragraph{Size Loss.} The size loss encourages the model to select fewer nodes by penalizing the sum of node probabilities. For a batch of graphs, we compute:
\[
\mathcal{L}_{\text{size}} = \frac{1}{B}\sum_{j=1}^{B}\frac{1}{n_j}\sum_{i=1}^{n_j}p_i^{(j)}
\]
where $B$ is the batch size, $n_j$ is the number of nodes in graph $j$, and $p_i^{(j)}$ is the probability for node $i$ in graph $j$. Normalization by graph size ensures that the loss scales appropriately across graphs of different sizes.

\paragraph{Coverage Loss.} The coverage loss penalizes configurations where vertices are not adequately dominated. For each vertex $u$, we compute the probability that it remains uncovered (neither in the dominating set nor adjacent to any node in it). Let $N[u] = \{u\} \cup \{v : (u,v) \in E\}$ denote the closed neighborhood of $u$. Assuming that the inclusion of nodes in the dominating set are independent events across the closed neighborhood $N[u]$, the probability that $u$ is not dominated is:
\[
q_u = \prod_{v \in N[u]}(1-p_v)
\]

To maintain numerical stability, we compute this in log-space:
\[
\log q_u = \sum_{v \in N[u]}\log(1-p_v + \epsilon)
\]
where $\epsilon = 10^{-4}$ is added for numerical stability. The coverage loss aggregates these probabilities:
\[
\mathcal{L}_{\text{coverage}} = \frac{1}{B}\sum_{j=1}^{B}\sum_{u \in V_j}q_u
\]

This formulation provides a smooth, differentiable approximation to the discrete constraint that every vertex must be dominated.

\paragraph{Equalizer Loss.} To encourage the model to make decisive predictions (probabilities close to 0 or 1 rather than hovering around 0.5), we introduce an equalizer term that penalizes uncertainty:
\[
\mathcal{L}_{\text{equalizer}} = \frac{1}{B}\sum_{j=1}^{B}\frac{1}{n_j}\sum_{i=1}^{n_j}p_i^{(j)}(1-p_i^{(j)})
\]

This term is maximized when $p_i = 0.5$ and minimized when $p_i \in \{0,1\}$, thus encouraging binary-like decisions. The relative weights of these loss components are set empirically: $\alpha = 70$ for size penalty, $\beta = 2$ for coverage penalty, and $\gamma = 50$ for the equalizer term. These coefficients balance the competing objectives and are tuned based on validation performance.

\subsection{Multi-Strategy Decoding Algorithms}

During inference, we convert the continuous node probabilities into discrete dominating sets using four distinct decoding strategies. For each graph instance, we execute all four algorithms and return the smallest of the four resulting dominating sets. This ensemble approach ensures robust performance across diverse graph structures, as different strategies excel on different graph topologies.

\subsubsection{Three-Phase Strategy}

This algorithm combines statistical outlier detection, probabilistic sampling, and greedy completion in three sequential phases:

\textbf{Phase 1: Outlier Detection.} We identify nodes with exceptionally high probabilities using the interquartile range (IQR) method. For the probability distribution, we compute $Q_1$, $Q_3$, and $\text{IQR} = Q_3 - Q_1$. Nodes with probabilities exceeding $Q_3 + 1.5 \times \text{IQR}$ are immediately added to the dominating set and marked as covering themselves and their neighbors.

\textbf{Phase 2: Probabilistic Selection.} For nodes that remain uncovered after Phase 1, we perform stochastic selection by comparing each node's probability $p_i$ against a uniform random threshold $t_i \sim \text{Uniform}(0,1)$. Nodes satisfying $p_i > t_i$ are added to the dominating set, and their neighborhoods are marked as covered.

\textbf{Phase 3: Greedy Completion.} If any nodes remain uncovered after the first two phases, we perform a greedy pass: iterating through remaining nodes in descending order of probability and adding each node that covers at least one previously uncovered vertex. This phase guarantees that the final solution is a valid dominating set.

\subsubsection{Prune-Greedy Strategy}

This strategy employs an initial greedy selection followed by two refinement passes to improve solution quality:

\textbf{Initial Greedy Selection.} Nodes are processed in descending probability order. For each node, if it is not yet covered, it is added to the dominating set and all its neighbors are marked as covered.

\textbf{Leaf Refinement.} After the initial selection, we identify all leaf nodes (degree-1 vertices) that were included in the dominating set. For each such leaf, we replace it with its unique neighbor if the neighbor is not already in the dominating set. This exploits the observation that in an optimal dominating set, it is always preferable to include a leaf's neighbor rather than the leaf itself, as the neighbor can potentially cover additional nodes.

\textbf{Redundancy Pruning.} Finally, we examine nodes in ascending probability order (checking low-confidence selections first) and attempt to remove each node from the dominating set. If removal does not violate the domination property (i.e., all vertices remain covered), the node is permanently removed. This eliminates redundant selections that may have occurred during the greedy phase.

\subsubsection{Threshold-Greedy Strategy}

This deterministic approach uses a two-phase strategy:

\textbf{Threshold Selection.} All nodes with probabilities exceeding a fixed threshold $\tau = 0.9$ are immediately added to the dominating set, and their neighborhoods are marked as covered.

\textbf{Greedy Completion.} For any vertices that remain uncovered after threshold selection, we perform a greedy completion pass, processing nodes in descending probability order and adding each node that covers at least one uncovered vertex.

This strategy is particularly effective for graphs where the model produces highly confident predictions, as it exploits high-probability nodes while ensuring complete coverage through the completion phase.

\subsubsection{Neighbor-Greedy Strategy}

This variant implements an enhanced greedy strategy that can select already-covered nodes if doing so provides additional coverage benefits:

Nodes are processed in descending probability order. For each node, we check whether it is already covered and whether all its neighbors are covered. If either condition is false (i.e., the node or some of its neighbors are uncovered), the node is added to the dominating set. This allows the algorithm to select covered nodes that can help dominate uncovered neighbors, providing greater flexibility than standard greedy approaches that skip covered nodes entirely.

\subsection{Ensemble Selection}

For each test instance, all four decoding algorithms are executed independently, and we return the smallest of the four resulting sets, $D_{\text{pred}} = \arg\min_{k \in \{1,\dots,4\}} |D_k|$. Because every decoding strategy terminates only once all vertices are dominated, each $D_k$ is a valid dominating set and therefore satisfies $|D_k| \geq |D_{\text{opt}}|$. The absolute error $|D_k| - |D_{\text{opt}}|$ is thus a strictly increasing function of $|D_k|$ for a fixed instance, so selecting the smallest set is equivalent to selecting the set of lowest error, while requiring no knowledge of $|D_{\text{opt}}|$ at inference time. This ensemble approach leverages the complementary strengths of each strategy: the Three-Phase Strategy excels on medium-density graphs with clear probability distributions, Prune-Greedy performs well on sparse graphs where refinement eliminates unnecessary selections, Threshold-Greedy succeeds when predictions are highly confident, and Neighbor-Greedy handles cases where flexible node selection improves coverage efficiency.

\subsection{Training Procedure}

We train our model using the Adam optimizer with learning rate $\eta = 5 \times 10^{-3}$ and no weight decay. Gradient clipping with maximum norm 1.0 is applied to stabilize training. Models are trained for 800 epochs with batch size 16 on a dataset of 12,000 connected graphs with sizes $n \in [10, 40]$. 

The training data consists of a mixture of graph types designed to expose the model to diverse structural patterns: very sparse trees (15\%), medium-density Erdős-Rényi graphs with $p = 1.5\log n / n$ (70\%), dense Erdős-Rényi graphs with $p = 0.35$ (10\%), and near-complete graphs with approximately 2\% of edges randomly removed (5\%). All graphs are generated to be connected by adding cross-edges between components when necessary. Ground-truth minimum dominating sets are computed using OR-Tools CP-SAT solver for exact optimization and are used solely for post-hoc evaluation; they are never provided to the model as supervision during training. See Section~\ref{sec:experimental} for complete details.

During training, we monitor all three loss components as well as solution quality metrics on a validation set. Training typically converges within 400-600 epochs, after which the loss components stabilize and solution quality plateaus. The entire training procedure for a single model takes approximately 2-3 hours on a single NVIDIA GPU.

\section{Applications and Architectural Considerations}\label{sec:applications}

\subsection{Real-World Applications of Dominating Set}

The Minimum Dominating Set problem has diverse practical applications across multiple domains, with particularly significant implications for social network analysis.

\subsubsection{Social Network Applications}

In the era of online social networks, the MDS problem and its variants address critical challenges in information diffusion and influence maximization. \textbf{Viral marketing} represents one of the most commercially significant applications~\cite{kempe2003influence}: companies seek to identify a minimal set of influential individuals (seed users) who, when given promotional content or product samples, will trigger cascading adoption throughout their social circles. This word-of-mouth mechanism is significantly more effective and trusted than traditional advertising channels.

The Positive Influence Dominating Set (PIDS) variant~\cite{wang2009pids} enhances this model by requiring that each person have a sufficient fraction of their friends (typically half) in the influence set before adopting a behavior or product. This threshold-based model more accurately reflects real social influence dynamics, where individuals are more likely to adopt behaviors when multiple friends have already done so. Applications extend beyond marketing to \textbf{public health interventions}, where identifying key individuals for behavioral change programs (smoking cessation, healthy lifestyle adoption, vaccination campaigns) can create positive cascades throughout communities.

\textbf{Political campaigns} leverage dominating set concepts to identify key voters or community leaders whose endorsement can reach entire neighborhoods or demographic groups. \textbf{Information dissemination} strategies for emergency alerts, public safety announcements, or rumor blocking benefit from minimal dominating set identification to ensure rapid, complete coverage while minimizing resource expenditure.

Scale-free social networks, which exhibit power-law degree distributions, present unique computational challenges. Research has shown that dominating sets in these networks tend to be larger than in random graphs~\cite{molnar2013minimum}, making efficient heuristics essential. The presence of highly connected hub nodes creates both opportunities (hubs can reach many people) and challenges (requiring more seeds to satisfy threshold conditions for high-degree nodes in PIDS formulations).

\subsubsection{Other Application Domains}

Beyond social networks, dominating sets solve practical problems in \textbf{wireless sensor networks}~\cite{cone2019}, where they identify minimal placements for relay nodes or base stations ensuring network-wide coverage while reducing deployment costs. In \textbf{power grid monitoring}~\cite{haynes2002power}, dominating sets determine optimal locations for monitoring devices that can detect faults throughout electrical networks. For \textbf{biological networks}~\cite{milenkovic2011dominating}, identifying dominating sets helps locate critical proteins or genes whose regulation affects entire functional modules. The connected dominating set variant is particularly important for mobile ad-hoc networks (MANETs), where dominating nodes form a virtual backbone for efficient routing~\cite{butenko2003cds,wan2003cds}.

\subsection{Advantages and Limitations of Our Approach}

Our GNN-based framework offers several key advantages. The unsupervised training paradigm eliminates the computational expense of computing optimal solutions for large training datasets, making the approach practical for real-world deployment where generating ground-truth labels for thousands of graph instances would be prohibitive. The learned heuristics demonstrate strong generalization to unseen graph distributions and larger problem sizes, suggesting that the model captures fundamental structural patterns rather than memorizing training instances. The probabilistic formulation provides a smooth optimization landscape and naturally handles the discrete constraints of the MDS problem through differentiable approximations.

However, our approach has limitations. The independence assumption in the coverage loss, while providing efficient gradients, may be less accurate for extremely dense graphs or highly correlated neighborhood structures. Performance may degrade on graphs with very high average degree (approaching $O(n)$), where the independence assumption underlying the probabilistic approximation becomes less reliable. The ensemble decoding strategy, while improving solution quality, increases inference time compared to single-pass greedy algorithms. For latency-critical applications, this trade-off between quality and speed must be carefully considered.

\subsection{Architectural Design Choices}

Two common concerns in deep GNN architectures---oversmoothing and oversquashing~\cite{alon2021bottleneck}---are not significant issues for our approach due to several design factors. Our model uses only two GNN layers, which is intentionally shallow. For the dominating set problem, the critical information is primarily local: whether a node should be in the dominating set depends mainly on its immediate neighborhood structure and degree. With two layers, each node's representation is influenced by its 2-hop neighborhood, providing sufficient context without the information loss associated with deeper architectures.

The MDS problem exhibits strong locality properties, aligning well with shallow GNN designs. We explicitly concatenate degree information with learned embeddings before the final prediction layer, preserving crucial structural information that might otherwise be smoothed away. This degree-augmented architecture effectively implements a form of skip connection that maintains discriminative power across diverse graph structures.

Our experimental results empirically validate these choices: the model successfully distinguishes between nodes and produces diverse probability distributions across graphs of varying sizes and densities, confirming that node representations retain sufficient discriminative power. Recent theoretical work has shown that for problem classes with local structure, shallow GNNs can achieve competitive performance while avoiding the pitfalls of deeper architectures~\cite{neuman2022bandlimited}, and our results align with these findings.

\section{Experiments and Results}\label{sec:experimental}

\subsection{Experimental Setup}

\subsubsection{Dataset Generation}

We generate a training dataset of 12,000 connected graphs using a stratified mixture designed to expose the model to diverse graph structures. Graphs have between 10 and 40 nodes, with the distribution mixing four structural types:

\begin{itemize}
\item \textbf{Trees (15\%):} Very sparse connected graphs with exactly $n-1$ edges, generated using random spanning tree algorithms. These represent the sparsest possible connected structure.

\item \textbf{Medium-density ER (70\%):} Erdős-Rényi random graphs with edge probability $p = 1.5 \log n / n$, a standard choice that produces connected graphs with moderate density. Components are connected by adding random cross-edges when necessary.

\item \textbf{Dense ER (10\%):} Erdős-Rényi graphs with higher edge probability $p = 0.35$, representing denser network structures typical of tightly-knit communities or clustered systems.

\item \textbf{Near-complete (5\%):} Nearly complete graphs generated by removing approximately 2\% of edges from a complete graph, simulating highly connected systems with sparse missing links.
\end{itemize}

For each graph instance, we compute the exact minimum dominating set using the OR-Tools CP-SAT solver, which formulates MDS as a constraint programming problem with Boolean variables and domination constraints. These ground-truth solutions are used solely for evaluation; they are never provided to the model during training. The dataset is split into training (80\%), validation (10\%), and test (10\%) sets with fixed random seeds to ensure reproducibility.

\subsubsection{Evaluation Protocol}

Performance is assessed using relative error, defined as:
\[
\text{RelError} = \frac{|D_{\text{pred}}| - |D_{\text{opt}}|}{|D_{\text{opt}}|}
\]
where $D_{\text{pred}}$ is the predicted dominating set and $D_{\text{opt}}$ is the optimal solution. We also measure inference time (milliseconds per graph) to assess computational efficiency. All experiments use fixed random seeds to ensure reproducibility.

For each test instance, we execute all four decoding algorithms (3-Phase, Prune-Greedy, Threshold-Greedy, and Neighbor-Greedy) and report the smallest of the four resulting dominating sets; as shown in Section~\ref{sec:methodology}, this requires no knowledge of the optimal solution. This ensemble strategy ensures robust performance by leveraging the complementary strengths of each decoding approach.

\subsubsection{Baselines and Evaluation}

We compare our GNN-based GP approach against classical heuristics, an 
unsupervised GNN baseline that ablates our core architectural and training 
contributions, and a supervised learning-based method specifically designed 
for MDS.
\\
\textbf{Classical Baselines:}
\begin{itemize}

\item \textbf{Trivial-Greedy:} NetworkX's built-in \texttt{dominating\_set()} 
function~\cite{networkx}, which selects nodes in arbitrary graph-traversal order 
with no degree or coverage awareness. Included as a lower-bound reference.

\item \textbf{Static-Greedy:} A classical greedy heuristic that sorts nodes by 
degree once at initialization and iteratively selects the highest-degree uncovered 
node until all vertices are dominated. Achieves an $O(\ln \Delta)$-approximation 
guarantee but uses no dynamic feedback during selection.

\item \textbf{Dynamic-Greedy:} An improved greedy variant that re-evaluates each 
candidate node's marginal coverage gain after every selection step, always picking 
the node that dominates the most currently uncovered vertices. Strictly dominates 
Static-Greedy at the cost of $O(n^2)$ recomputation per step. Note that while this 
method and our \emph{Neighbor-Greedy} decoder share a similar intuition, they differ 
fundamentally: Dynamic-Greedy operates on raw graph structure alone, whereas 
Neighbor-Greedy uses GNN-learned node probabilities to guide selection order.

\item \textbf{Simulated Annealing (SA):} A metaheuristic initialized from the 
Static-Greedy solution, performing 5{,}000 swap and removal moves with exponential 
temperature cooling from $T_0=2.0$ to $T_f=0.01$. Infeasible candidates are 
repaired greedily before acceptance evaluation. Provides strong solution quality 
at the cost of high per-instance computation time.

\end{itemize}

\textbf{Learning-Based Baseline:}
\begin{itemize}

\item \textbf{EGN-MDS (Karalias \& Loukas~\cite{karalias2020erdos}):} 
An adaptation of the Erdős Goes Neural unsupervised framework for MDS. 
A GCN is trained with a probabilistic loss combining a size penalty 
($\alpha{=}20$) and a coverage penalty ($\beta{=}10$), without an 
equalizer term or ensemble decoding. Solutions are decoded using the 
Method of Conditional Expectations~\cite{karalias2020erdos}. This 
baseline directly ablates our two core contributions: removing the 
equalizer yields uncertain probability distributions that the decoder 
cannot resolve cleanly, and removing the ensemble leaves a single 
decoding path. Note that the original EGN codebase targets Max Clique 
and Min Cut; the MDS loss was derived and implemented following the 
formulation in~\cite{karalias2020erdos}.

\item \textbf{Kothapalli et al.\ (GCN-MDS):} A supervised GCN-based heuristic 
specifically designed for MDS~\cite{kothapalli2023mds}. A two-layer GCN scores 
nodes for inclusion and these scores are integrated into an iterative greedy 
procedure (IG-GCN). Unlike our approach, this method requires pre-computed 
ground-truth dominating sets as supervision during training.

\end{itemize}

We also investigated S2V-DQN~\cite{dai2017learning}, a reinforcement 
learning approach that constructs solutions sequentially via Q-learning. 
However, the official implementation requires compiling a custom C++ 
graph library with CUDA~\cite{dai2017learning}, which is incompatible 
with our experimental setup. A pure-Python reimplementation failed to 
converge within the computational budget available (3{,}000 training 
episodes), consistently producing near-complete dominating sets. We 
therefore exclude S2V-DQN from the quantitative comparison and note 
this as a limitation of our evaluation.

Similarly, we did not include GCON~\cite{wenkel2025gcon} as a direct 
baseline, as that work evaluates on different benchmark datasets and 
graph scales, making direct numerical comparison uninformative. We 
note that GCON also addresses MDS in an unsupervised setting and 
represents a strong complementary approach; a systematic comparison 
on shared benchmarks is left for future work.

\subsection{Qualitative Analysis: Graph Visualization}

\begin{figure*}[!t]
\centering
\includegraphics[width=\textwidth]{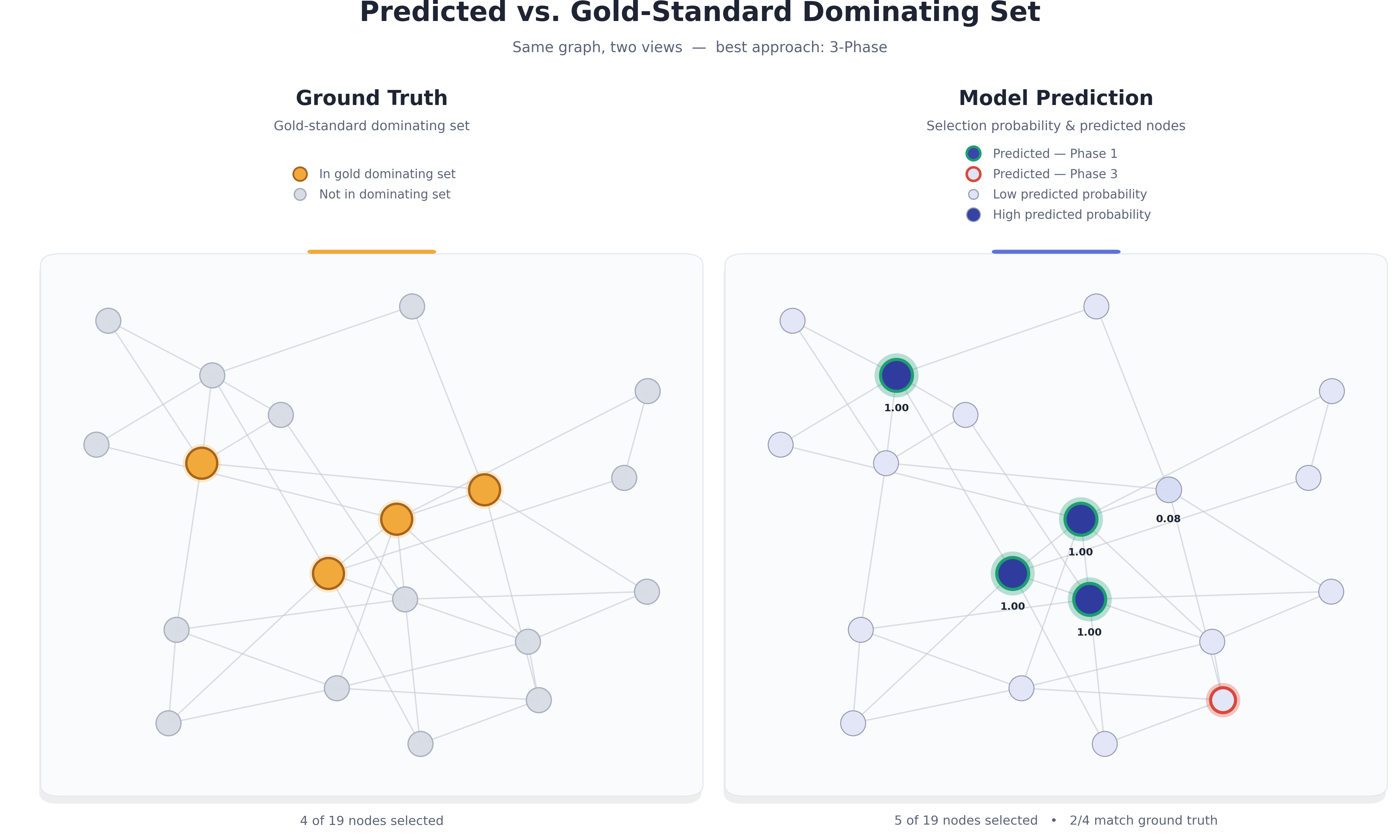}
\caption{Ground truth versus model prediction on a representative 19-node test 
graph, using the 3-Phase decoding strategy. Left: the optimal dominating set 
(orange, 4 of 19 nodes). Right: predicted selection probabilities, with node 
fill shaded from light (low probability) to dark navy (high probability); a 
green outline marks the four nodes selected during Phase 1 (outlier detection) 
and a red outline marks the one additional node added during Phase 3 (greedy 
completion), giving a 5-node predicted set. Two of the four optimal nodes are 
recovered exactly (2/4 match), and the predicted set fully dominates all 19 
vertices.}
\label{fig:graph_examples}
\end{figure*}

Figure~\ref{fig:graph_examples} presents a representative 19-node test graph, 
comparing the gold-standard dominating set against our model's predicted 
selection probabilities under the best-performing decoder for this instance, 
the 3-Phase strategy.

For this instance, the 3-Phase decoder assigns probability $p_i = 1.00$ to four 
nodes during Phase 1 outlier detection (green outlines), all of which are added 
directly to the dominating set. One additional node is selected during Phase 3 
greedy completion (red outline) to cover the vertex left undominated after 
Phase 1, yielding a 5-node predicted set against an optimal size of 4. Two of 
the four optimal nodes are recovered exactly; the remaining two optimal nodes 
are substituted with alternative, equally valid covering nodes elsewhere in 
the graph, since multiple minimum dominating sets of the same size can exist 
for a single instance. Non-selected nodes generally receive low predicted 
probabilities (e.g.\ 0.08 for the node shown in the figure), confirming that 
the model's outputs are decisive rather than diffuse.

This example illustrates two complementary aspects of our approach: the 
learned probabilities are highly confident wherever the model identifies a 
clear candidate, and the multi-phase decoding procedure reliably closes the 
small coverage gaps that high-confidence predictions alone leave 
behind---here adding only a single node beyond the optimal solution size.

\subsection{Performance on Benchmark Graphs}

Tables~\ref{tab:benchmarks-classical}--\ref{tab:runtimes-learned} compare all
methods on the seven benchmark instances (split into classical and
learning-based method groups for readability). In
Tables~\ref{tab:benchmarks-classical} and~\ref{tab:benchmarks-learned}, bold
entries indicate that the method achieved the optimal dominating set size.

These seven graphs are established small-scale benchmarks from the social 
network analysis literature: the bottlenose dolphin association 
network~\cite{lusseau2003dolphins}, Padgett's Florentine families marriage 
and business network~\cite{padgett1993robust}, Krackhardt's high-tech company 
managers network~\cite{krackhardt1987cogn}, Coleman's high school friendship 
network~\cite{coleman1964intro}, the Les Mis\'{e}rables character 
co-occurrence network~\cite{knuth1993graphbase}, Knecht's classroom 
friendship network (Klas12b)~\cite{knecht2008friendship}, and the Japanese 
macaque dominance interaction network~\cite{takahata1991diachronic}.\footnote{Several 
of these graphs are originally distributed as directed, weighted, and/or 
multiplex networks (e.g.\ multiple relation types or directional ties). In 
each such case we construct a single simple undirected graph for evaluation 
by taking the union of all layers/relations, symmetrizing directed ties so 
that an edge $(u,v)$ is retained whenever either direction is recorded, and 
discarding edge weights.}

\begin{table*}[!t]
\centering
\footnotesize
\setlength{\tabcolsep}{5pt}
\caption{DS size on benchmark instances: classical baselines. Bold = optimal
solution achieved. Bottom row reports average relative error across all
graphs for these methods.}
\label{tab:benchmarks-classical}
\begin{tabular}{lccccc}
\toprule
\textbf{Graph} & \textbf{$|V|$} & \textbf{Opt}
  & \textbf{Trivial-G} & \textbf{Static-G} & \textbf{Dynamic-G} \\
\midrule
dolphins            & 62 & 14 & 25 & 19 & 16 \\
florentine families & 16 &  4 &  7 & \textbf{4} & \textbf{4} \\
high tech company   & 21 &  1 &  3 & \textbf{1} & \textbf{1} \\
highschool          & 70 & 11 & 19 & 14 & 12 \\
Les Mis\'{e}rables  & 77 & 10 & 22 & 23 & \textbf{10} \\
Klas12b             & 26 &  5 &  8 &  7 &  6 \\
macaques            & 62 &  2 &  6 & \textbf{2} & \textbf{2} \\
\midrule
Avg.\ rel.\ error (\%) & & & 115 & 33 & 6 \\
\bottomrule
\end{tabular}
\end{table*}

\begin{table*}[!t]
\centering
\footnotesize
\setlength{\tabcolsep}{5pt}
\caption{DS size on benchmark instances: metaheuristic and learning-based
methods. Bold = optimal solution achieved. Bottom row reports average
relative error across all graphs for these methods.}
\label{tab:benchmarks-learned}
\begin{tabular}{lccccc}
\toprule
\textbf{Graph} & \textbf{$|V|$} & \textbf{Sim-Ann}
  & \textbf{EGN-MDS} & \textbf{Kothapalli} & \textbf{Ours} \\
\midrule
dolphins            & 62 & \textbf{14} & 31 & \textbf{14} & 16 \\
florentine families & 16 & \textbf{4}  & 14 & \textbf{4}  & \textbf{4} \\
high tech company   & 21 & \textbf{1}  &  2 & \textbf{1}  & \textbf{1} \\
highschool          & 70 & \textbf{11} & 23 & \textbf{11} & 14 \\
Les Mis\'{e}rables  & 77 & \textbf{10} & 36 & \textbf{10} & 11 \\
Klas12b             & 26 & \textbf{5}  & 15 & \textbf{5}  & \textbf{5} \\
macaques            & 62 & \textbf{2}  &  5 & \textbf{2}  & \textbf{2} \\
\midrule
Avg.\ rel.\ error (\%) & & 0 & 170 & 0 & 7 \\
\bottomrule
\end{tabular}
\end{table*}

\begin{table*}[!t]
\centering
\footnotesize
\setlength{\tabcolsep}{5pt}
\caption{Inference time (ms) per graph instance: classical baselines. SA
runs 5{,}000 iterations per graph.}
\label{tab:runtimes-classical}
\begin{tabular}{lcccc}
\toprule
\textbf{Graph} & \textbf{$|V|$}
  & \textbf{Trivial-G} & \textbf{Static-G} & \textbf{Dynamic-G} \\
\midrule
dolphins            & 62 & 0.07 & 0.09 &  6.05 \\
florentine families & 16 & 0.02 & 0.02 &  0.10 \\
high tech company   & 21 & 0.01 & 0.02 &  0.04 \\
highschool          & 70 & 0.09 & 0.08 &  0.89 \\
Les Mis\'{e}rables  & 77 & 0.03 & 0.06 &  0.72 \\
Klas12b             & 26 & 0.03 & 0.05 &  0.35 \\
macaques            & 62 & 0.02 & 0.06 &  0.43 \\
\bottomrule
\end{tabular}
\end{table*}

\begin{table*}[!t]
\centering
\footnotesize
\setlength{\tabcolsep}{5pt}
\caption{Inference time (ms) per graph instance: metaheuristic and
learning-based methods. GNN times include single forward pass and
ensemble decoding. Bold = fastest among methods with $\leq 10\%$ average
relative error.}
\label{tab:runtimes-learned}
\begin{tabular}{lccccc}
\toprule
\textbf{Graph} & \textbf{$|V|$}
  & \textbf{Sim-Ann} & \textbf{EGN-MDS} & \textbf{Kothapalli} & \textbf{Ours} \\
\midrule
dolphins            & 62 & 1017.7 &  53.6 & 405 & \textbf{41.3} \\
florentine families & 16 &  226.5 &  29.6 &  52 & \textbf{20.9} \\
high tech company   & 21 &  407.2 &   9.1 &  83 & \textbf{15.4} \\
highschool          & 70 & 1123.8 &  72.1 & 618 & \textbf{44.7} \\
Les Mis\'{e}rables  & 77 & 1116.9 &  90.7 & 456 & \textbf{46.0} \\
Klas12b             & 26 &  445.4 &  11.8 & 106 & \textbf{27.9} \\
macaques            & 62 & 1615.7 &  59.6 & 422 & \textbf{29.2} \\
\bottomrule
\end{tabular}
\end{table*}

Dynamic-Greedy achieves a slightly lower average relative error than our 
method (6\% versus 7\%) and is also faster across every benchmark instance in 
this study, including the two largest graphs tested ($|V|=70$ and $|V|=77$). 
This reflects the small scale of the available real-world benchmarks: at 
$n \leq 77$, the $O(n^2)$ recomputation cost of Dynamic-Greedy remains 
negligible in absolute terms. Whether our single-pass GNN yields a practical 
speed advantage over Dynamic-Greedy would need to be verified on graphs 
substantially larger than those in our current benchmark suite.

Simulated Annealing achieves optimal solutions on all seven instances but at 
a cost of 226--1616~ms per graph, making it 10--55$\times$ slower than our 
method and entirely impractical for real-time or large-scale deployment. 
Kothapalli et al.\ also achieve optimal on all instances but require supervised 
training on pre-labeled data and are 2--14$\times$ slower at inference than 
our approach. Among all methods achieving $\leq 10\%$ average relative error 
--- Dynamic-Greedy, SA, Kothapalli, and ours --- our method is the only one 
that is simultaneously unsupervised, single-pass at inference, and under 50~ms 
on all tested instances. The EGN-MDS baseline, which shares our unsupervised paradigm but omits 
the equalizer term and ensemble decoding, achieves 170\% average relative 
error --- worse than classical greedy methods. This confirms that neither 
the GCN architecture nor the probabilistic loss formulation alone is 
sufficient: the equalizer term is essential for producing binary-like 
probability distributions that decoders can resolve into small dominating 
sets, and the ensemble is necessary to exploit those distributions 
effectively.

\section{Discussion}

Our experimental results demonstrate that graph neural networks can effectively learn heuristics for the Minimum Dominating Set problem through unsupervised training. The probabilistic loss function successfully balances competing objectives of minimizing set size and ensuring complete coverage, as evidenced by the convergent training dynamics and low relative errors achieved on test instances.

The strong performance on benchmark graphs indicates that patterns learned from synthetic training data transfer effectively to real-world networks with diverse structural properties. This generalization capability is particularly valuable for practical applications where training data may consist of easily-generated synthetic graphs, but deployment targets real-world social networks, infrastructure graphs, or biological networks.

The unsupervised training paradigm offers significant practical advantages. Computing optimal dominating sets for large graphs using exact solvers is computationally expensive, often requiring hours or days for instances with hundreds of nodes. Our approach eliminates this bottleneck during training, requiring ground-truth solutions only for evaluation purposes. This makes the framework particularly suitable for domains where optimal solutions are prohibitive to compute, such as large-scale social network analysis or real-time network optimization.

The applicability to social network influence maximization is particularly promising. Our method can identify small sets of influential individuals for viral marketing campaigns, with potential adaptations to handle threshold-based influence models (such as PIDS) through modifications to the coverage loss. The practical inference times (under 50 ms per graph for benchmark instances) make the approach viable for interactive applications where campaign strategies must be computed rapidly.

However, limitations remain. As noted in Section~\ref{sec:applications}, performance may degrade on extremely dense graphs where the independence assumption in the coverage loss becomes less accurate. Additionally, while our model generalizes well to graphs within the training size range and moderately larger instances, performance on very large graphs (thousands of nodes) remains to be systematically evaluated. We expect this to require scaling the training distribution as well as the architecture: our model was trained exclusively on graphs with $n \in [10, 40]$, and structural properties that matter at scale---such as power-law degree distributions, community structure, and long-range dependencies---are largely absent from that regime. Retraining on a distribution that includes substantially larger and more heterogeneous graphs is therefore a prerequisite for strong large-graph performance, independent of any architectural change. Future work could additionally investigate architectural modifications (such as hierarchical message-passing or graph coarsening techniques) to improve scalability.

\section{Conclusion}

We have presented a graph neural network framework for the Minimum Dominating Set problem that achieves strong performance through unsupervised training. Our key contributions include: (1) a novel probabilistic loss function with explicit size, coverage, and equalizer terms that eliminates the need for ground-truth solutions during training; (2) a comprehensive ensemble of four decoding strategies that leverages complementary algorithmic strengths to ensure robust performance across diverse graph structures; (3) empirical demonstration of effective generalization from synthetic training graphs to real-world benchmark instances; and (4) validation of practical applicability through competitive performance and reasonable inference times.

Our results demonstrate that neural approaches to combinatorial optimization can be both practical and effective for NP-hard graph problems with significant real-world applications. The unsupervised training paradigm makes learned heuristics accessible for domains where optimal solutions are computationally prohibitive, opening opportunities for deployment in large-scale applications including viral marketing, public health interventions, and information dissemination in social networks.

The comparison with classical baselines reveals a clear quality-speed frontier 
among existing approaches. Trivial-Greedy and Static-Greedy are fast but poor 
(33--115\% average error). Dynamic-Greedy closes most of this gap (6\% error) 
but scales quadratically and cannot leverage learned structure. Simulated 
Annealing solves the quality problem but at a computational cost that grows 
with iterations, not graph structure. Our approach occupies a unique position 
on this frontier: by amortizing the cost of learning across thousands of 
training graphs, inference reduces to a single forward pass regardless of graph 
size, achieving near-identical average error to Dynamic-Greedy (7\% vs.\ 6\%). 
On the present benchmarks Dynamic-Greedy remains the faster of the two, and 
establishing a practical speed advantage for our method requires graphs 
substantially larger than those tested here; against the supervised Kothapalli 
baseline, however, our approach is 2--14$\times$ faster at inference while 
requiring no labelled training data.

Comparison with the supervised baseline of Kothapalli et al.~\cite{kothapalli2023mds} 
highlights a fundamental trade-off between solution quality and training accessibility. 
Their method achieves optimal solutions on all seven benchmark graphs by leveraging 
labeled training data, while our unsupervised approach matches optimal on four of seven 
graphs and comes within one to three nodes on the remaining three. For practitioners, 
this gap may be an acceptable cost: computing exact optimal dominating sets for thousands 
of training graphs using exact solvers is computationally prohibitive at scale, whereas 
our method requires no such labeling budget. Furthermore, the inference-time advantage 
of our approach (under 50~ms versus up to 618~ms) is practically significant for 
real-time or interactive applications such as live campaign targeting on social networks.

The failure of EGN-MDS further isolates our contributions: given the 
same unsupervised training paradigm and graph scale, our equalizer term 
and ensemble decoding account for the entire gap between 170\% and 7\% 
average relative error.

Future directions include extending the framework to related variants such as positive influence dominating set (incorporating threshold-based influence dynamics) and connected dominating set (ensuring the selected nodes form a connected subgraph). Investigating transfer learning scenarios where models pre-trained on synthetic graphs are fine-tuned on small amounts of real-world data could bridge the gap between synthetic and real distributions. Architectural enhancements such as hierarchical message-passing or attention-based aggregation mechanisms may improve performance on very large graphs. Finally, theoretical analysis of approximation guarantees achievable by learned heuristics would provide valuable insights into the fundamental capabilities and limitations of GNN-based approaches for dominating set and related problems.

\backmatter

\section*{Declarations}

\subsection*{Availability of data and materials}
The datasets generated during the current study will be made available
upon reasonable request. The implementation code will be made available
upon publication.

\subsection*{Competing interests}
The authors declare that they have no competing interests.

\subsection*{Funding}
The authors received no specific funding for this work.

\subsection*{Authors' contributions}
E.A.\ conceived the study, developed the methodology, implemented the
model, conducted experiments, and wrote the manuscript. B.B.\ supervised
the research, contributed to the methodology design, and reviewed and
edited the manuscript. M.S.\ contributed to exploratory work in the
early stages of the project. All authors read and approved the final
manuscript.

\subsection*{Acknowledgements}
Not applicable.

\bibliography{refs}

\end{document}